\documentclass[10pt,twocolumn,letterpaper]{article}

\usepackage{iccv}
\usepackage{times}
\usepackage{epsfig}
\usepackage{graphicx}
\usepackage{amsmath}
\usepackage{amssymb}
\usepackage{enumitem}
\setlist{nolistsep}

\usepackage{caption}
\usepackage{flushend}

\usepackage{subcaption}

\usepackage[pagebackref=true,breaklinks=true,letterpaper=true,colorlinks,bookmarks=false]{hyperref}

\def\realnumbers{\mathbb{R}}
\newcommand*{\red}{\textcolor{red}}
\newcommand*{\green}{\textcolor{green}}
\newcommand*{\blue}{\textcolor{blue}}

\def\etal{\emph{et al.}}
\def\ws{webly-supervised }

\def\r2d{R(2+1)D }

\iccvfinalcopy 

\def\iccvPaperID{4132} 

\ificcvfinal\fi
\begin{document}

\title{Image to Video Domain Adaptation Using Web Supervision}

\author{Andrew Kae\\
{\tt\small andrew.kae@gmail.com}
\and
Yale Song\\
Microsoft Cognition\\
{\tt\small yalesong@microsoft.com}
}

\maketitle


\begin{abstract}
Training deep neural networks typically requires large amounts of labeled data which may be scarce or expensive to obtain for a particular target domain. As an alternative, we can leverage \emph{webly-supervised} data (i.e. results from a public search engine) which are relatively plentiful but may contain noisy results. 
In this work, we propose a novel two-stage approach to learn a video classifier using \ws data. We argue that learning appearance features and then temporal features sequentially, rather than simultaneously, is an easier optimization for this task. We show this by first learning an image model from web images, which is used to initialize and train a video model. 
Our model applies domain adaptation to account for potential domain shift present between the source domain (\ws data) and target domain and also accounts for noise by adding a novel attention component.
We report results competitive with state-of-the-art for \ws approaches on UCF-101 (while simplifying the training process) and also evaluate on Kinetics for comparison.
\end{abstract}

%
%
%
%
%
%
%


\section{Introduction}
\label{sec:intro}
Action recognition in videos is a well-studied problem in computer vision with many important applications in areas such as surveillance, search, and human-computer interaction.
Training deep neural networks typically requires a large labeled dataset. However, it may be difficult to obtain enough labeled data because it may be too scarce or too expensive to obtain.
%
We can instead leverage \emph{webly-supervised} data (i.e. results from a public search engine) which are relatively plentiful but may be noisy. 

\begin{figure}
\centering
\includegraphics[width=0.45\textwidth]{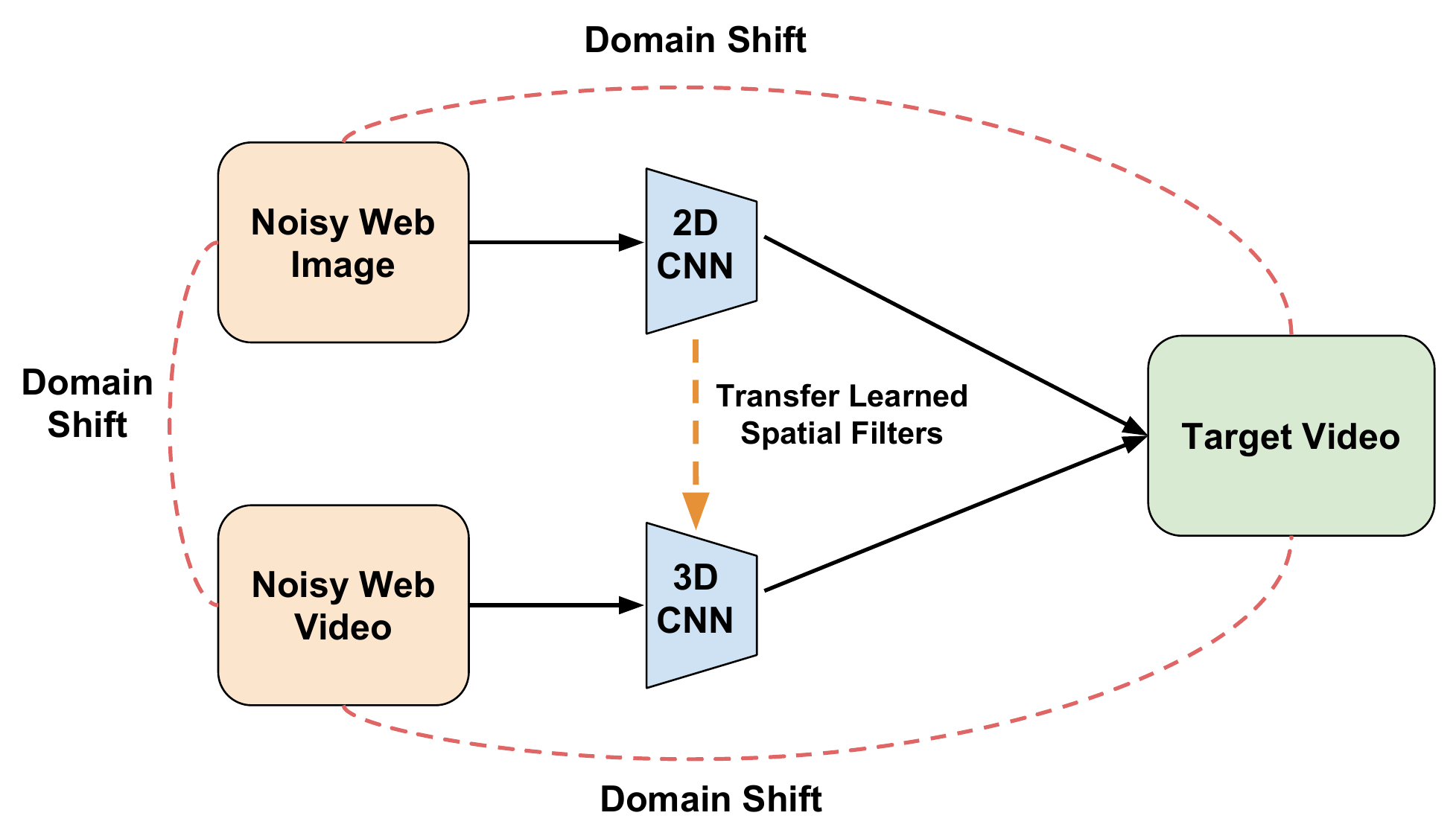}
\caption{Given \ws images and videos (source domains), we learn a video classifier for the target domain. The model is learned in a two-stage process by 1) learning an image model (2D-CNN) and 2) transferring the spatial filters to the video model (3D-CNN) to continue training. The model also accounts for domain shift and noise present in the \ws data.}
\label{fig:high_level}
\end{figure}
%
The high-level overview of our model is shown in Figure~\ref{fig:high_level}. %
The noisy web image and web video domains are considered source domains that we want to domain adapt into the target domain.  
We present a two-stage approach to first learn an image model using a 2D-CNN, transfer the learned spatial weights to a 3D-CNN, and continue training a video model.
Since our goal is to learn a video classifier, we can potentially learn from web videos only, but we argue that our proposed two-stage process is more appropriate for learning from noisy, \ws data. Web videos are likely to be noisier than web images since web videos typically contain many frames that are irrelevant to the target concept.
Thus it may be easier to learn spatial features first, based on the relatively cleaner web images, and then learn the temporal features afterward.  
Previous work~\cite{r2d} has also hypothesized that it may be difficult to learn both spatial and temporal features simultaneously.
%
We present empirical results in Section~\ref{sec:experiments} showing that our two-stage process, which separates learning appearance and temporal features, outperforms a model that learns both jointly.

\begin{figure*}[ht!]
  \centering
  \begin{subfigure}[b]{0.24\textwidth}
      \includegraphics[width=\textwidth]{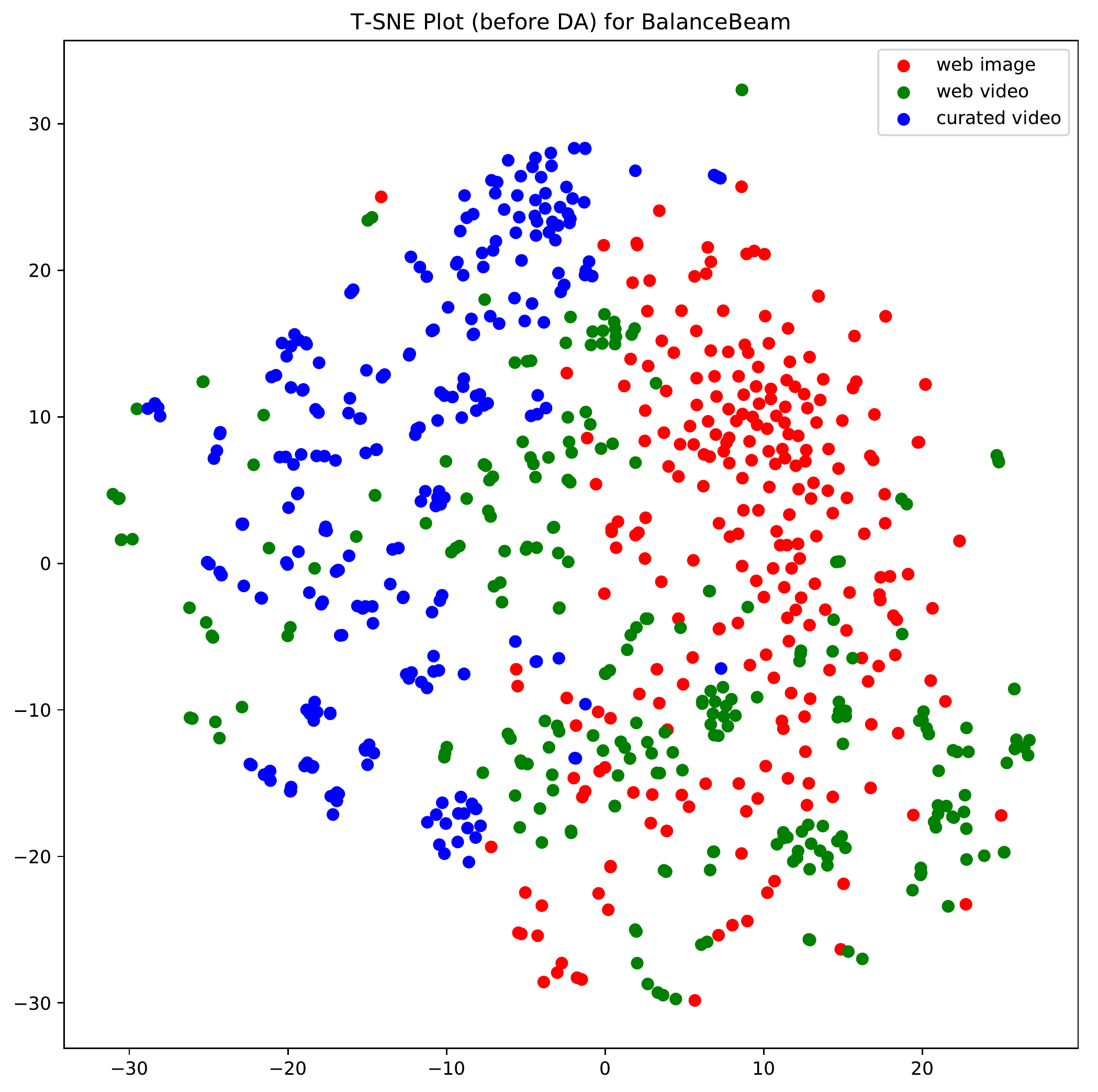}
  \end{subfigure}
  \begin{subfigure}[b]{0.24\textwidth}
      \includegraphics[width=\textwidth]{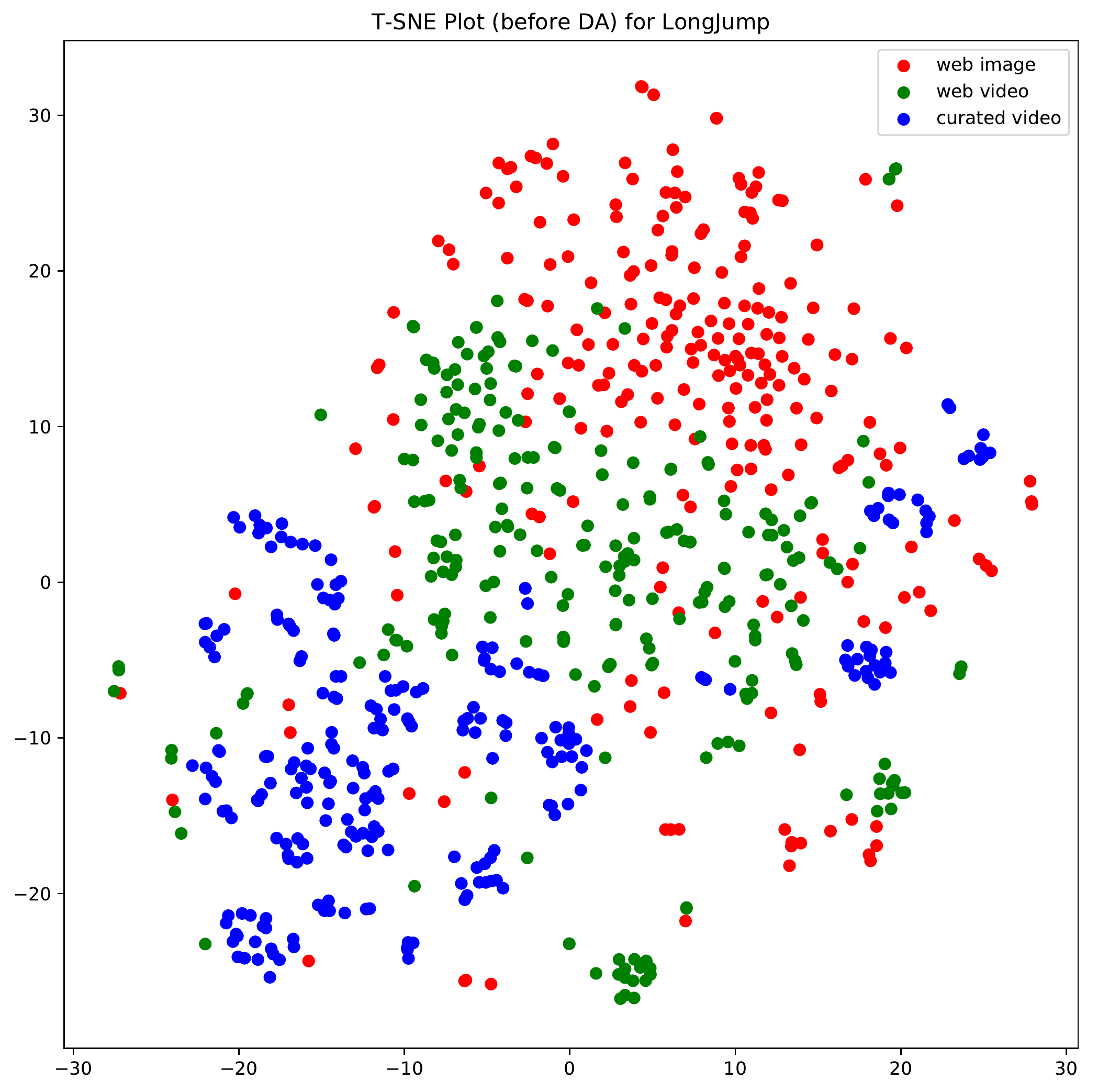}
  \end{subfigure}
  \begin{subfigure}[b]{0.24\textwidth}
      \includegraphics[width=\textwidth]{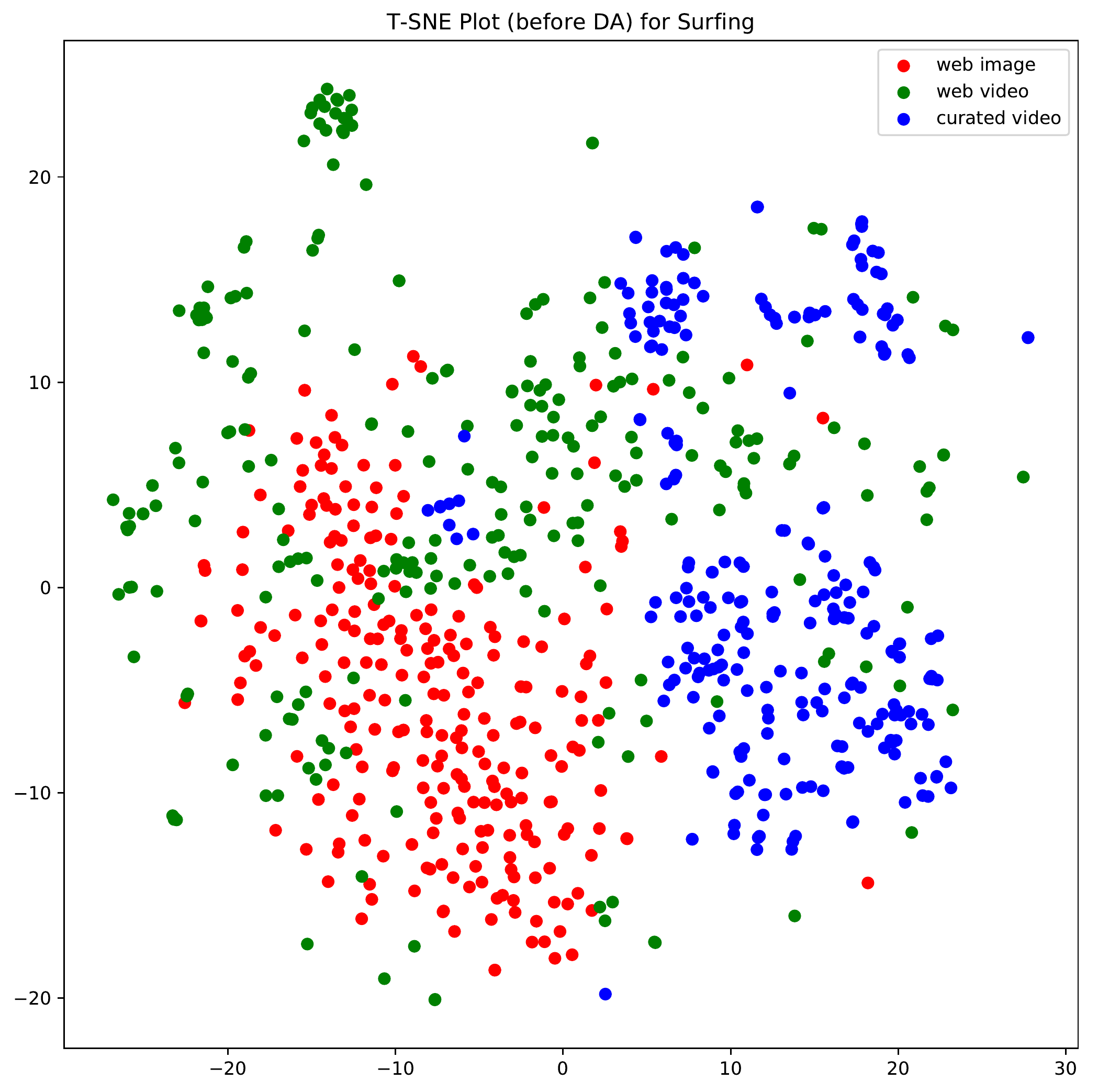}
  \end{subfigure}
  \begin{subfigure}[b]{0.24\textwidth}
      \includegraphics[width=\textwidth]{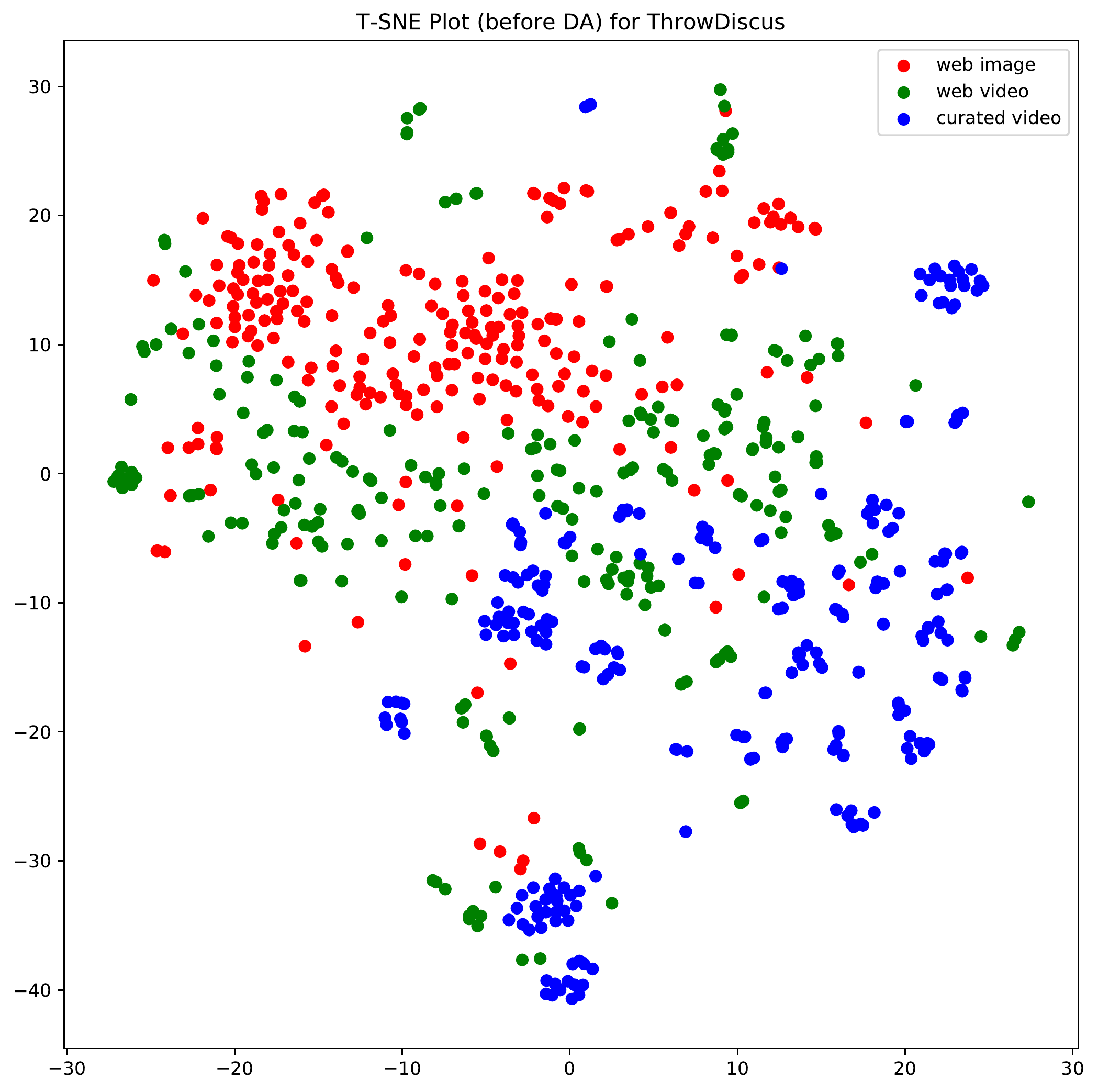}
  \end{subfigure}      
  \\
  \begin{subfigure}[b]{0.24\textwidth}
      \includegraphics[width=\textwidth]{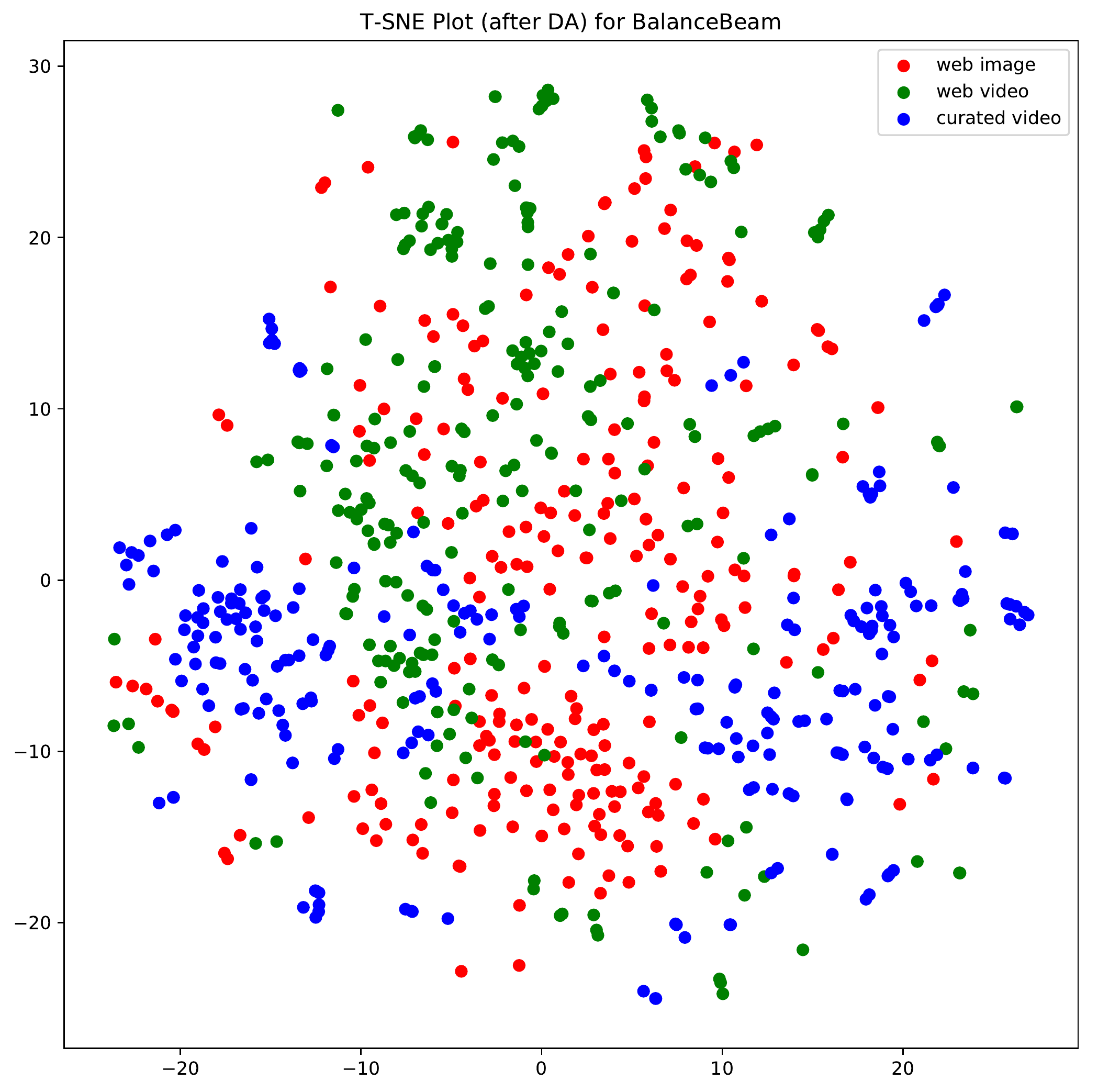}
  \end{subfigure}
  \begin{subfigure}[b]{0.24\textwidth}
      \includegraphics[width=\textwidth]{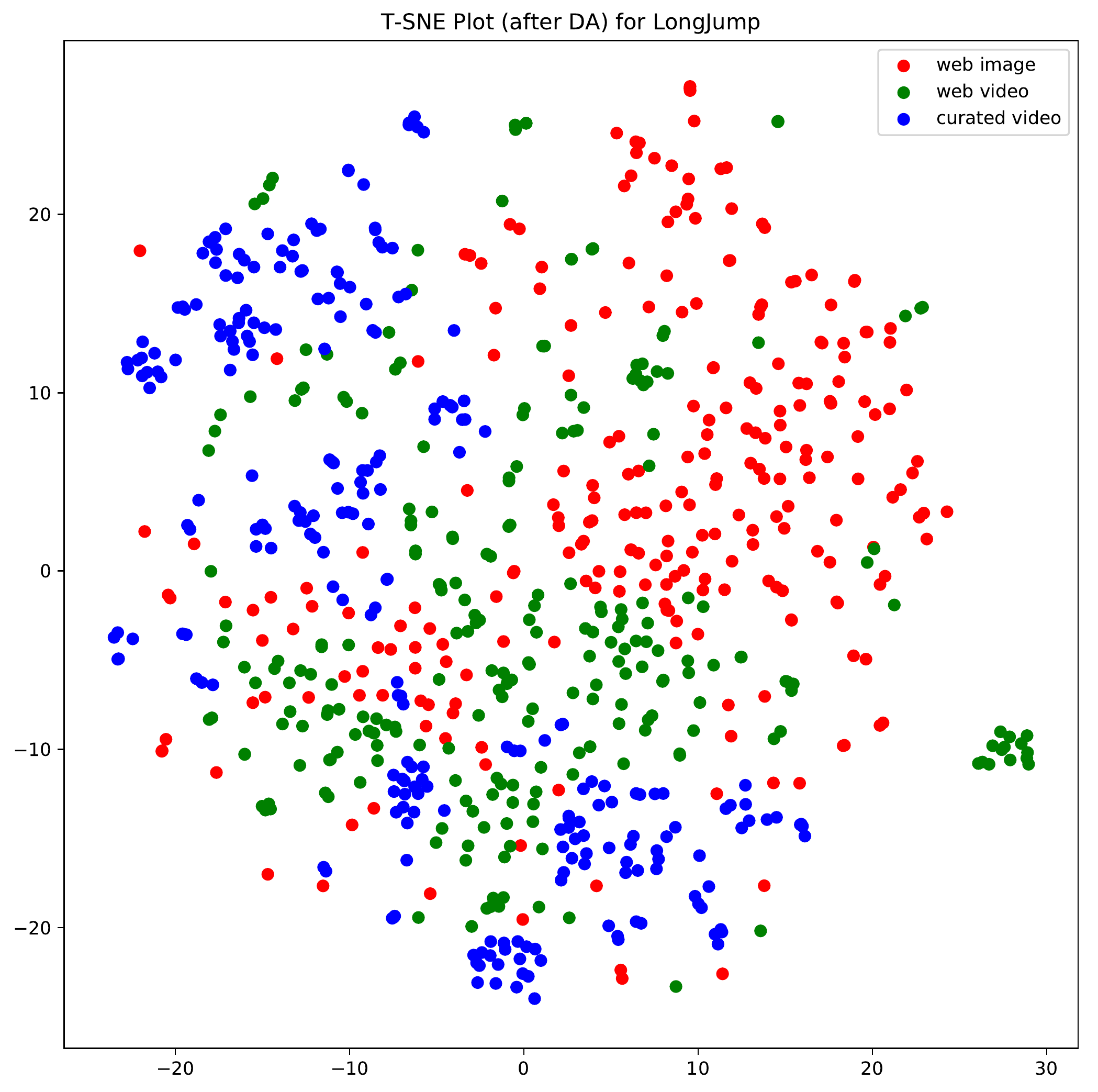}
  \end{subfigure}
  \begin{subfigure}[b]{0.24\textwidth}
      \includegraphics[width=\textwidth]{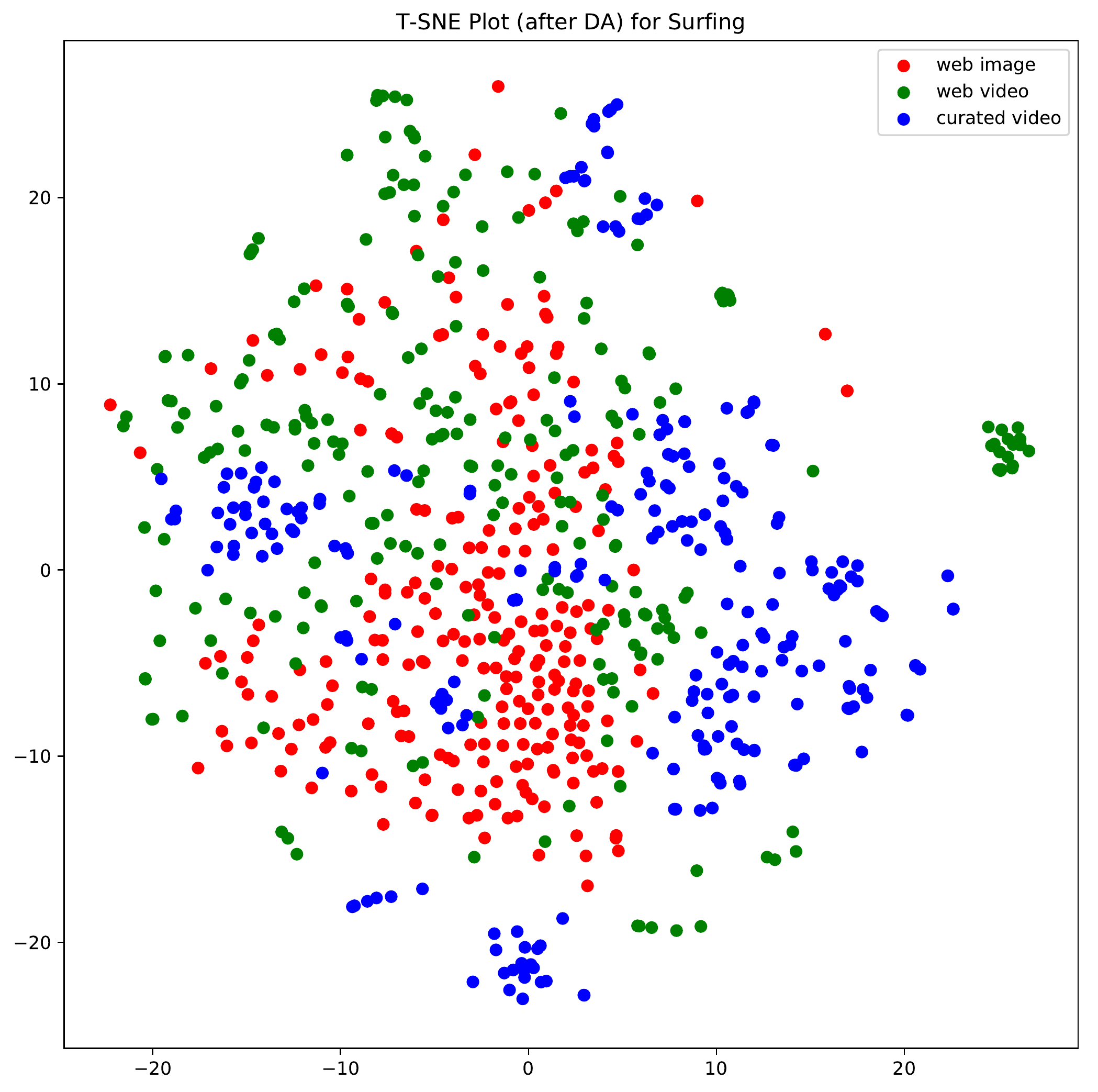}
  \end{subfigure}
  \begin{subfigure}[b]{0.24\textwidth}
      \includegraphics[width=\textwidth]{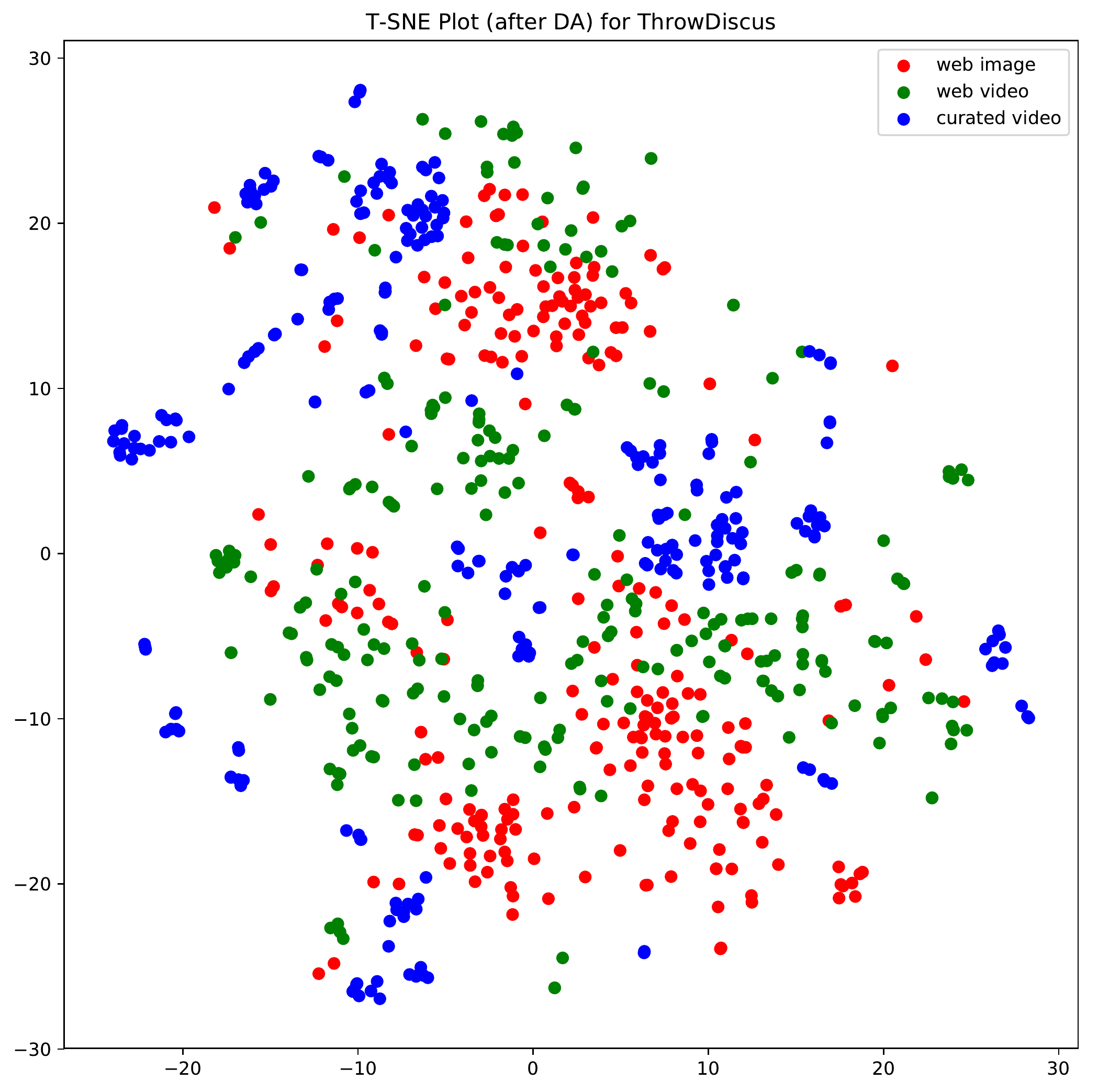}
  \end{subfigure}      
  \caption{\textbf{T-SNE Plots.} We randomly sampled from the web image (\textbf{\red{red}} points), web video (\textbf{\green{green}} points) and target video (\textbf{\blue{blue}} points) (UCF-101~\cite{ucf-101}) domains and show the T-SNE~\cite{tsne} plots of 4 actions: balance beam, long jump, surfing, and throw discus.  The first row contains the T-SNE plot before domain adaptation using pre-trained RN-34~\cite{resnet} and the second row shows the same actions after the network has been domain adapted.  Plot best viewed in color.}
  \label{fig:tsne}
\end{figure*}

In addition to the challenges of learning the appearance and motion, there are two additional issues with training on \ws data. First, there is potential domain shift between the different domains. For example, comparing web images and videos, many web images are typically high-resolution and shot with high-quality cameras, while web videos are typically lower resolution and may contain motion blur and other artifacts. Second, there may be noise present in \ws data that may degrade performance.  
For example \ws data may contain either the wrong concept entirely or a mix of relevant and irrelevant concepts (i.e. only a subset of frames in a video may correspond to the target concept).


To account for domain shift, domain adaptation has been successfully used for tasks such as mapping from MNIST~\cite{mnist} to StreetView digits~\cite{adda, dann}, RGB to depth images~\cite{adda} and webcam to product images~\cite{dann}. 
%
In our work we incorporate an adversarial training component taken from Generative Adversarial Networks (GAN)~\cite{gan}. 
To account for the noise present in \ws data, we incorporate a novel attention component to reduce the effect of irrelevant examples, inspired by attention models for machine translation~\cite{attention}.
%

In this work, the target domain consists of \textit{curated} videos, containing only a single concept or activity.  We consider these curated videos to be a separate domain from web images and web videos. We assume there are relatively few irrelevant chunks from videos in the target domain compared to web videos. For example, this setting may be appropriate if the target domain was surveillance videos. 
%

To check whether there is indeed a difference between the separate domains, we extracted embeddings from random images/frames from each domain using ResNet-34~\cite{resnet} and visualized T-SNE~\cite{tsne} plots for four different action categories from UCF-101~\cite{ucf-101}: Balance Beam, Long Jump, Surfing, Throw Discus.  
The top row corresponds to the embeddings before domain adaptation (DA) for curated video frames (\textbf{\blue{blue}} points), web video frames (\textbf{\green{green}} points), and web images (\textbf{\red{red}} points).  The bottom row corresponds to the embeddings after our DA (detail in Section~\ref{sec:model}).  
In the top row, before DA, there are visibly distinct regions corresponding to the three domains of web images, web videos and curated videos (we used UCF-101~\cite{ucf-101} videos), which may indicate domain differences.
After DA, the different domains are packed closer together.

To summarize, our contributions include:
\begin{itemize}
\item A novel two-stage approach to first learn spatial weights from a 2D-CNN and then transfer these weights to a 3D-CNN to learn temporal weights.
\item A novel attention component to account for noise present in \ws data.
\item Results competitive with state-of-the-art on UCF101~\cite{ucf-101}, while simplifying training.
\end{itemize}

\section{Related Work} \label{relatedWork}

\textbf{Webly-Supervised Learning}. 
Previous work using \ws data include~\cite{singh15zeroshot,chenEventDriven,doLessAchieveMore,UntrimmedNets}.  Gan \etal~\cite{webly} jointly match images and frames in a pre-processing step before using a classifier while LeadExceed~\cite{leadExceed} uses multiple steps to filter out noisy images and frames. 
In contrast, our model does not have pre-processing steps and learns to downweight noisy images as part of model training.  

Li \etal~\cite{attentionTransfer} use web images to perform domain adaptation and learn a video classifier, but they manually filter out irrelevant web images beforehand whereas we incorporate this step into our model.

There has also been related work in using attention for weakly-supervised learning. Zhuang \etal~\cite{attendInGroups} stack noisy web image features together with the assumption that at least one of the images is correctly labeled. They then learn an attention model to focus on the correctly labeled images. 
%
UntrimmedNet~\cite{UntrimmedNets} generates clip proposals from untrimmed web videos and also incorporates an attention component for focusing on the proposals with the correct action.
In contrast, our model learns from both images and videos and ties attention closely with domain adaptation.



\textbf{Video Classification}. 3D-CNN video models such as C3D~\cite{c3d}, P3D~\cite{p3d}, I3D~\cite{quoVadis}, \r2d~\cite{r2d} are appealing for video classification since they learn appearance and motion features jointly.  
%
I3D~\cite{quoVadis} uses full 3D filters, while \r2d~\cite{r2d} and P3D~\cite{p3d} decompose the spatio-temporal convolution into a spatial convolution followed by a temporal convolution.  
The design of our 3D-CNN is partly inspired by these latter approaches because of this elegant decomposition, which allows us to reuse spatial filters from a conventional 2D CNN. We could potentially use the same bootstrapping technique to inflate 2D to 3D filters as in I3D~\cite{quoVadis}, but initializing and fixing the 2D filters may allow for easier training (more detail in Section~\ref{sec:model}).

%

\textbf{Domain Adaptation}. 
%
%
There has been much work in adapting GANs~\cite{gan} for domain adaptation. Models such as PixelDA~\cite{pixelda} learn to generate realistic-looking samples from the source distribution, while others such as DANN~\cite{dann} learn a domain-invariant feature representation. We adopt this latter approach in our work. 
%
%
Other related works include Adversarial Discriminative Domain Adaptation (ADDA)~\cite{adda} which learns a piecewise model by pre-training a classifier on the source domain and then adds the adversarial component later. 
Tzeng et. al~\cite{simultaneousDeep} learn domain-invariance by incorporating a domain confusion loss (similar to a discriminator loss) and transferring class correlations between domains to preserve class-specific information.
%
%
%
Luo \etal~\cite{luo_nips17_label} propose a similar model to ours but for the supervised setting, 
and add a semantic-transfer loss to encourage transfer of class-specific information.

The main difference between our model and these approaches is that we use \ws data and assume the source and target domains may contain noisy labels, which is a considerably more difficult yet practical scenario.
Lastly there is recent work by Zhang \etal~\cite{importanceWeighted} that is similar to our model in that they also have a domain-adversarial component and perform instance weighting to account for noise in the source data.  However they use a dual-discriminator approach for instance weighting whereas we use an attention-based component.  In addition our model is designed specifically for image to video domain adaptation and classification.

%

%


%

\section{Model} \label{sec:model}
Our goal is to learn a video classifier in the target domain by training on the \ws (source) image and video domains.  
We propose a two-stage approach by first learning an image model using a standard 2D-CNN, transferring the learned spatial weights to a 3D-CNN and then continuing training on videos. We learn a separate model for images and videos since it may be difficult to learn appearance and motion features simultaneously.  

Our model should: (1) learn appearance features in the image model and motion features in the video model (2) transfer the learned spatial weights from the image model to the video model properly (3) account for noise present in the \ws images and videos and (4) perform domain adaptation from the \ws domain to the target domain.

%
%
%

\begin{figure}[tp]
\centering
\includegraphics[width=0.45\textwidth]{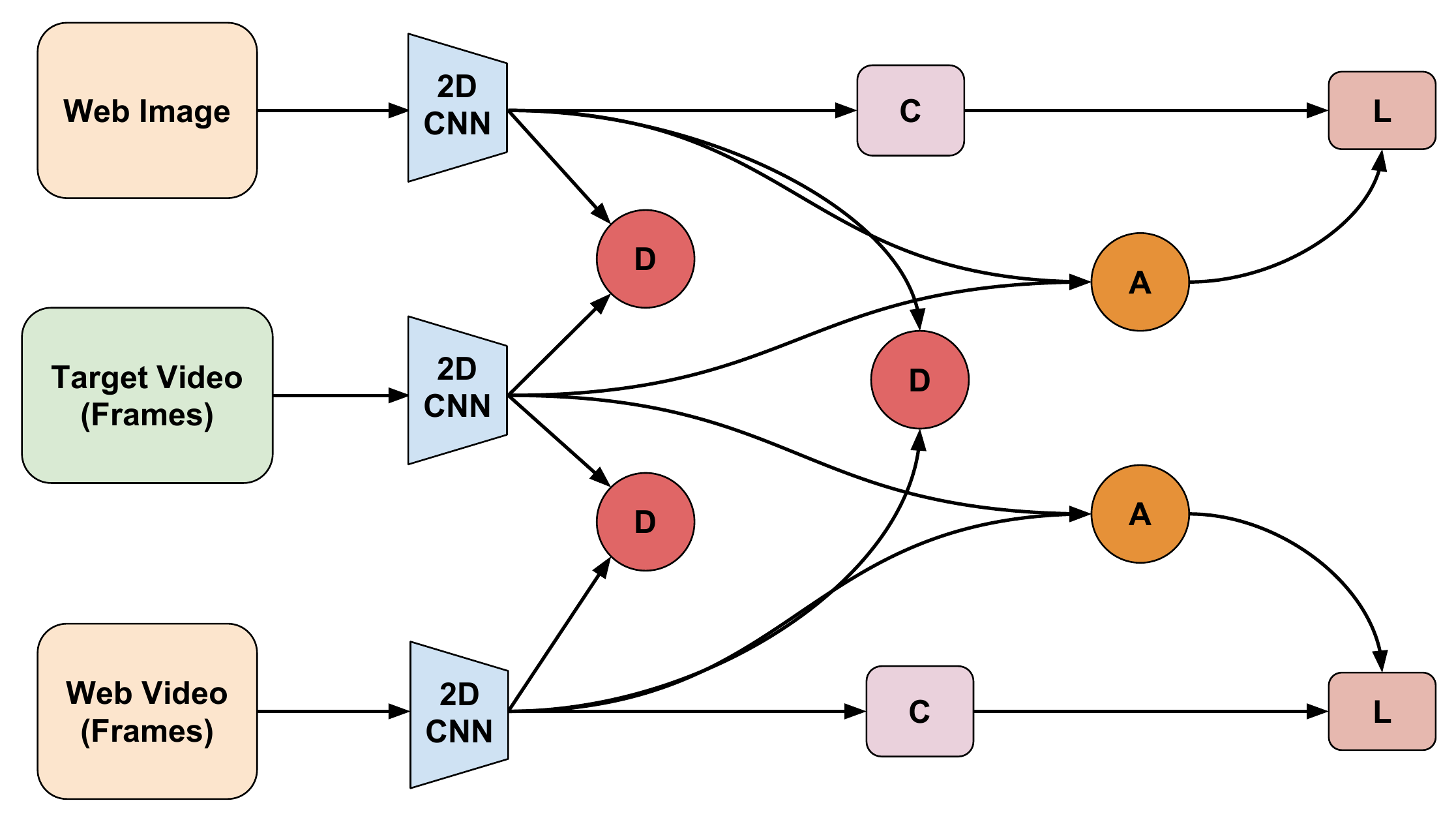}
\caption{\textbf{Image Model}.   Triplet network with branches corresponding to web images, web video frames and target video frames.  We add discriminators $D$ to enforce domain invariance between the separate domains and add attention components $A$ to downweight irrelevant examples.  
$C$ corresponds to the classifiers and $L$ corresponds to the losses.}
\label{fig:image_model}
\end{figure}

%
The image model shown in Figure~\ref{fig:image_model} is a triplet network that performs both domain adaptation and attention-based filtering of noisy images.
The three branches correspond to web images, web video frames, and target video frames (without labels).  The image model learns domain invariance between the different domains and 
also uses an attention component to downweight irrelevant web images and web video frames, with respect to the target video frames.  
Intuitively we would like to downweight web images/frames that look different from target video frames.

\begin{figure}[tp]
\centering
\includegraphics[width=0.45\textwidth]{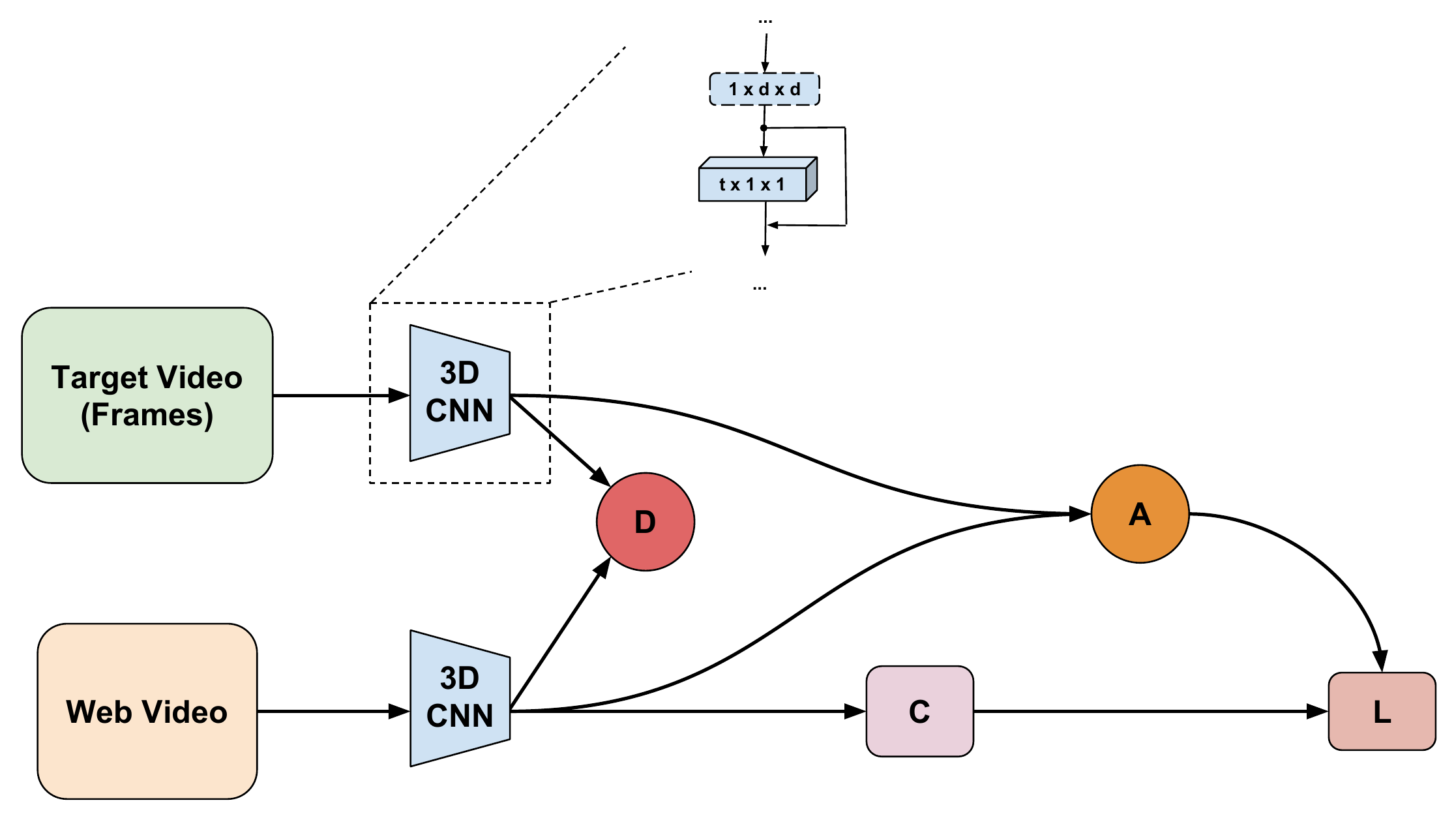}
\caption{\textbf{Video Model}. We use a Siamese model with branches corresponding to web video and curated video chunks.  We initialize the spatial weights in the 3D-CNN and add an attention component $A$ to reduce the noise from irrelevant shots or incorrect labels. $C$ corresponds to the classifier and $L$ corresponds to the loss.}
\label{fig:video_model}
\end{figure}

The video model shown in Figure~\ref{fig:video_model} is a Siamese network with branches corresponding to web videos and target videos (without labels).  Note that the inputs are now video chunks rather than images.
%
%
The spatial weights in the video model are initialized from the image model spatial weights and fixed (as indicated by the dashed lines in Figures~\ref{fig:video_model} and~\ref{fig:spatiotemporal_block}).  
Similar to the image model, the video model also contains domain adaptation and attention components.
%

%


\subsection{Notation}
Let us define following notation:
\begin{itemize}
\item $E$: encoder (either a 2D or 3D CNN) returns $\mathbb{R}^d$ 
\item $C$: classifier returns predictions among $L$ labels
\item $N^I, N^V, N^T$: number of webly-supervised image and videos, and curated (target) videos respectively
\item $\textbf{X}^I = \{ x_i^I, y_i^I \}_{i=1}^{N^I}$: set of webly-supervised images where $x_i^I$ is the $i$th image and $y_i^I$ is its corresponding label where $y_i^I \in \{1 .. L\}$ 
\item $\textbf{X}^V = \{ x_j^V, y_j^V \}_{j=1}^{N^V}$: set of webly-supervised videos where $x_j^V$ is the $j$th video and $y_j^V$ is its corresponding label and $y_j^V \in \{1 .. L\}$.  Each video $x_j^V$ consists of frames $\{ x_{jf}^V \}_{f=1}^{N_j^V}$ where $N_j^V$ is the number of frames in video $x_j^V$
\item $\textbf{X}^T = \{ x_k^T \}_{k=1}^{N^T}$: set of curated videos where $x_k^T$ is the $k$th video.  Each video $x_k^T$ consists of frames $\{ x_{kf}^T \}_{f=1}^{N_k^T}$ where $N_k^T$ is the number of frames in video $x_k^T$
\end{itemize}

\subsection{Classification}
We use ResNet-34~\cite{resnet} as the base architecture for both our image and video models, along with the standard softmax cross-entropy loss to train a classifier for both web images and web video frames. The losses are computed as 
\begin{align}
L_{wimage} &= \mathbb{E}_{x^I} \big[ - y^I \cdot \log(C(E(x^I))) \big] \nonumber \\
L_{wframe} &= \mathbb{E}_{x^V} \big[ - y^V \cdot \log(C(E(x^V))) \big] \nonumber
\end{align}
where the expectations are taken over examples $x^I$ and $x^V$ and $y^I, y^V$ are their corresponding webly-supervised labels. 
%


\subsection{Domain Adaptation}
%
Our goal is to learn an encoder $E$ that can produce feature embeddings that are indistinguishable between different domains. To this end, we use GANs~\cite{gan} as a way to perform domain adaptation. 
The discriminator $D$ tries to distinguish between embeddings generated from different domains (shown in Figure~\ref{fig:image_model}).   By optimizing over a mini-max objective, $E$ learns embeddings that can eventually ``fool'' $D$, thus learning a domain-invariant feature representation.
%
%

%
We define our domain-adaptation loss as
\begin{align} 
L^I &= \mathbb{E}_{x^T} \big[ \log D(E(x^T))   \big] + \mathbb{E}_{x^I} \big[  \log(1 - D(E(x^I)))  \big] \label{eq:image_loss} \\
L^V &= \mathbb{E}_{x^T} \big[ \log D(E(x^T))   \big] + \mathbb{E}_{x^V} \big[  \log(1 - D(E(x^V)))  \big] \label{eq:video_loss} \\
L^B &= \mathbb{E}_{x^I} \big[ \log D(E(x^I))   \big] + \mathbb{E}_{x^V} \big[  \log(1 - D(E(x^V)))  \big] \label{eq:both_loss} \\
L_{domain} &= L^I + L^V + L^B \label{eq:domain_loss}
\end{align}
Eqn.~\eqref{eq:image_loss} distinguishes between web images and target frames, Eqn.~\eqref{eq:video_loss} distinguishes between web video frames and target frames, and Eqn.~\eqref{eq:both_loss} distinguishes between web images and web video frames.  In each term, the first component corresponds to correctly distinguishing between different domains, and the second component tries to ``fool'' the discriminator $D$.
In addition, we use a multi-layer discriminator $D$ (similar to  ~\cite{luo_nips17_label}) structured as
\begin{align} 
d_l = D_l ( \sigma (d_{l-1} \oplus E_l(x)) ) \nonumber
\end{align} 
where $D_l$ refers to the discriminator at the $l$-th layer, $d_l$ refers to the discriminator output at the $l$-th layer, $\oplus$ denotes concatenation, $E_l(x)$ is the CNN embedding from the $l$-th layer, and $\sigma$ is the (relu) activation function.  
Intuitively we take the encoder outputs from multiple layers, concatenate them and feed them into a discriminator (a binary classifier).
We have empirically found this multi-layer discriminator to perform better than the single-layer version.

\subsection{Attention}
Learning from \ws data is difficult because it is inherently noisy.  For example, if we query for a term such as ``archery'', we may get some results containing the action of shooting a bow and arrow but we may also get advertisements for a sporting goods store, or product shots of archery equipment, which are likely less relevant for learning to recognize the action itself.  

We present a novel approach for filtering noisy data inspired by work from machine translation~\cite{attention}.  The attention component learns to ``filter out'' or downweight irrelevant images/frames by comparing the images and frames in each source domain batch to the images from the target domain batch.  Intuitively, images from the source domain batch that look very different to images in the target domain batch should be given low weight. For example, it is unlikely that an advertisement or product shot is going to look like frames from the target video which we assume is curated and contains only the action. In this way, we jointly learn the relevance of both web images and web videos by comparison to the target videos.
%
%
%

Note that this weighting is similar to the loss update from~\cite{importanceWeighted} but that is based on scores from a discriminator whereas our approach is based on learning a similarity function between the different domains.
%
Unlike previous work~\cite{webly} that performs pre-processing to filter out irrelevant images/frames, our approach learns a model of relevance jointly with other components during training.  In our approach we do not need to perform manual filtering or pre-processing and instead the filtering happens jointly with model training.
Zhuang et.al~\cite{attendInGroups} learns attention from stacking web image activations together.  In contrast our model learns attention through a comparison of the source and target domain, which may provide a more direct signal for inferring the attention weights. 

More formally, given a set of web images and their corresponding labels $\textbf{X}^I = \{ x_i^I, y_i^I \}_{i=1}^{N^I}$, we compute an attention score $\alpha_{i}$ for each image $x_i^I$ such that $\sum_{i}^{N^I} \alpha_{i} = 1$ (note that we compute these attention weights per batch during training).  
%
Let $E(x^I) \in \mathbb{R}^D$ denote the CNN embedding for a given image $x^I$.
%
We compute attention scores as follows
\begin{align} 
e_{ik} &= A(E(x_{i}^I), E(x_{k}^T)) \nonumber \\
&= E(x_{i}^I) \cdot W \cdot E(x_{k}^T)^\top \nonumber
\end{align}
where $e_{ik}$ indicates the similarity between web image $x_i^I$ and target image $x_k^T$ and $A$ is the attention model.  $W$ is a matrix with dimension $\realnumbers^{D \times D}$ and parameterizes the similarity between the embeddings from different domains.  The parameters for $W$ are learned along with the rest of the model parameters.  
We then compute
\begin{align} 
m_{it} = TopT(e_{i,1:N^{T}}) \nonumber
\end{align}
where $m_{it}$ consists of the top $T$ scores along the $i$th row.  In practice we observed better performance when summing over the top $T$ scores instead of all scores in the row. 
\begin{align} 
s_{i} &= \sum_{t=1}^T m_{it} \nonumber
\end{align}
We then compute the image attention weights as
\begin{align}
\alpha_{i}^I &= \frac{\exp(s_i / \tau)}{\sum_{j'=1}^N \exp(s_{i'} / \tau)} \nonumber \label{eq:attention}
\end{align}
where $\tau$ is a temperature term.  $\alpha_{i}^I$ is then used to weight the image $x_i^I$ in the cross-entropy loss.  The attention weights for video frames $\alpha_j^V$ are computed in the same way by comparing to the target video frames.

\subsection{Image Model}
The image model loss can be rewritten as:
\begin{align}
L_{wimage}' &= \mathbb{E}_{x^I} \big[ - \alpha^I \cdot y^I \cdot \log(C(E(x^I))) \big] \nonumber \\
L_{wframe}' &= \mathbb{E}_{x^V} \big[ - \alpha^V \cdot y^V \cdot \log(C(E(x^V))) \big] \nonumber \\
L_{image} &= L_{wimage}' + L_{wframe}'
\end{align}
where the $\alpha^I$ is used to weight the batch.
We also incorporate the domain adaptation loss from Equation~\ref{eq:domain_loss} to get a combined loss of
\begin{align} 
\min_{\theta_E, \theta_C} \max_{\theta_D} L_{image}(E, C) + \beta L_{domain}(E, D)
\end{align}
where $\beta$ is a tradeoff parameter between the weighted classification and domain adaptation terms.
In practice we use the \emph{gradient reversal layer}~\cite{dann} which multiplies the gradient from the discriminator by a negative constant during backpropagation, allowing us to perform optimization in one step instead of the usual two-step optimization for GANs.

\subsection{Video Model} 

\begin{figure}
\centering
\includegraphics[width=0.45\textwidth]{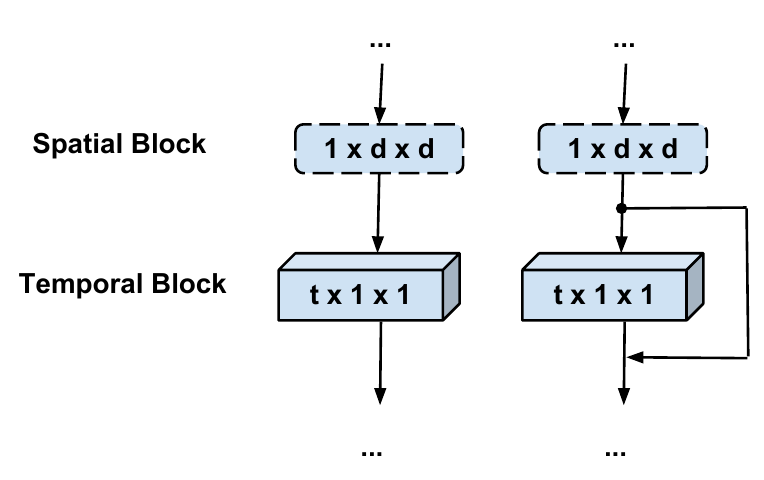}
\vspace{-0.5cm}
\caption{\textbf{Spatio-temporal Block.} (a) the decomposition of the spatiotemporal block into a 2D spatial filter followed by a 1D temporal filter (corresponds to \r2d~\cite{r2d} model)  (b)  our modified block with an added residual connection.  The spatial weights are initialized from the 2D CNN weights and fixed, as indicated by the dashed lines.}
\label{fig:spatiotemporal_block}
\end{figure}

The next step is to transfer the spatial filters learned from the image model to the video model.  We assume the spatial filters have been learned appropriately from the images and we want the video model to focus on learning the motion filters.  One natural way to capture this intuition is by \textit{sequentially} arranging the spatial filter followed by the temporal/motion filter, as shown in Figure~\ref{fig:spatiotemporal_block}a. 
In this way we elegantly decompose the spatiotemporal kernel into a spatial filter followed by a temporal filter.
Note that this formulation corresponds to the \r2d~\cite{r2d} architecture.


%
%
Unfortunately, there is a problem with simply placing the temporal block directly after the spatial block as shown in Figure~\ref{fig:spatiotemporal_block}a for our use case.
The ``good'' spatial filters (initialized from the image model) are now interleaved with untrained temporal filters, which means it is possible the output distribution of the spatiotemporal blocks can change significantly since the temporal filters still need to be learned (this is related to the problem of covariate shift~\cite{batchnorm} in training deep networks). We can reduce this effect somewhat by initializing all the temporal filters to the identity matrix, which will reduce the video model to the image model.
However, it is still possible that any slight change to the temporal weights may result in significant distribution changes to the spatiotemporal block, which can result in complicated optimization.  

Similar to the motivation of ResNet~\cite{resnet}, we propose to alleviate this issue by adding a residual connection (as shown in Figure~\ref{fig:spatiotemporal_block}b) and initializing the temporal filters to zero. This may be preferable to the previous approach since it may be easier to optimize to the residual.  
We have empirically found that we obtain better results by adding the residual connection, as detailed in Section~\ref{sec:experiments}.
The video model includes the same domain adaptation and attention components as earlier.  
The loss for the video domain in Figure~\ref{fig:video_model} is
\begin{align} 
L_{video}' &= \mathbb{E}_{x^V} \big[ - \alpha^V \cdot y^V \cdot \log(C(E(x^V))) \big]
\end{align}
which has the effect of ignoring or downweighting irrelevant video chunks in a soft way.  
%
%
The combined loss is similar to the image loss
\begin{align} 
\min_{\theta_E, \theta_C} \max_{\theta_D} L_{video}'(E, C) + \beta L_{domain}^V(E, D)
\end{align}
where $\beta$ is a tradeoff parameter.

\subsection{Training}
Putting all the pieces together, we first learn an image model (shown in Figure~\ref{fig:image_model}) using web images, web video frames, and target video frames as inputs.  Each input is fed into a 2D CNN where we extract embeddings that are used to compute the domain adaptation and weighted classification losses.  We then learn a video model (shown in Figure~\ref{fig:video_model}) by initializing the spatial filters from the 2D CNN and continue learning temporal filters from the videos.  Similar to the image model, we use a 3D CNN to extract embeddings which are used to compute the domain adaptation, and weighted classification losses.

\section{Experiments} \label{sec:experiments}

\subsection{Data}
We evaluate our model on a standard benchmark for video classification, UCF-101~\cite{ucf-101} and a larger dataset, Kinetics~\cite{kinetics}.  UCF-101 contains 101 action categories (such as ``golf swing'' or ``playing guitar'') consisting of about 13K video clips while Kinetics is a larger dataset containing 400 action categories and about 300K video clips\footnote{Note there is a more recent version of Kinetics with 600 classes but we use the older version with 400 classes due to computational limitations.}.
%
%
Similar to previous \ws approaches~\cite{webly, leadExceed}, we used standard image search engines (e.g. Bing, Google) to collect between 800-900 images (using the ``photo'' filter in the query) and YouTube to collect between 25-50 videos for each category.  
For our UCF experiments, the whole dataset consists of about 200K images and video keyframes.  We follow the same process for collecting \ws data for Kinetics and used about 400K images and video keyframes\footnote{We did not notice a significant improvement when using more data and we wanted to reduce computational overhead.}.
%

Since UCF and Kinetics videos are drawn from YouTube, it is possible that there may be some overlap with the \ws images and videos we collected.  To remove any potential overlap, we extracted CNN embeddings from keyframes from both the UCF/Kinetics videos and compared them to the web images and videos.  We then removed any web image or web video containing keyframes that had cosine similarity above a threshold (we used 0.9) with any UCF/Kinetics keyframe embedding.

\subsection{Implementation}
We used ResNet-34~\cite{resnet} 
as the base network for all experiments.  Every image is resized such that the shorter dimension is 256 and then a random crop of 224x224 is extracted.  For videos we first resize the video in a similar way and then use the Hecate~\cite{hecate} tool to extract keyframes and video chunks.  For each video chunk, we extract 24 frames, sampling every other frame to obtain a volume size of 12x224x224x3 per chunk.  We use a batch size of 32 for images and 10 for videos (note that we are limited by GPU memory in this case).
%
All models are coded in PyTorch~\cite{pytorch} and trained using stochastic gradient descent with momentum.
%
We use a held out validation set (20K \ws images and frames for UCF-101 and 40K for Kinetics) to choose model hyperparameters.
%
%
Both the 2D CNN encoders and classifiers in the image model (Figure~\ref{fig:image_model}) and the 3D CNN encoders and classifiers in the video model (Figure~\ref{fig:video_model}) have tied weights.

\subsection{Results}
Our initial hypothesis was there may be a domain difference between images and videos that may be reducing the effectiveness of the model.
To test this hypothesis, we trained a binary classifier (using ResNet-34~\cite{resnet}) to distinguish between web images and web video frames and found that the classifier was over 99\% accurate.  We hypothesize that compared to web images, web videos tend to be lower resolution and may contain motion blur and compression artifacts not typically found in web images.

\begin{figure*}
\centering
\includegraphics[width=\textwidth]{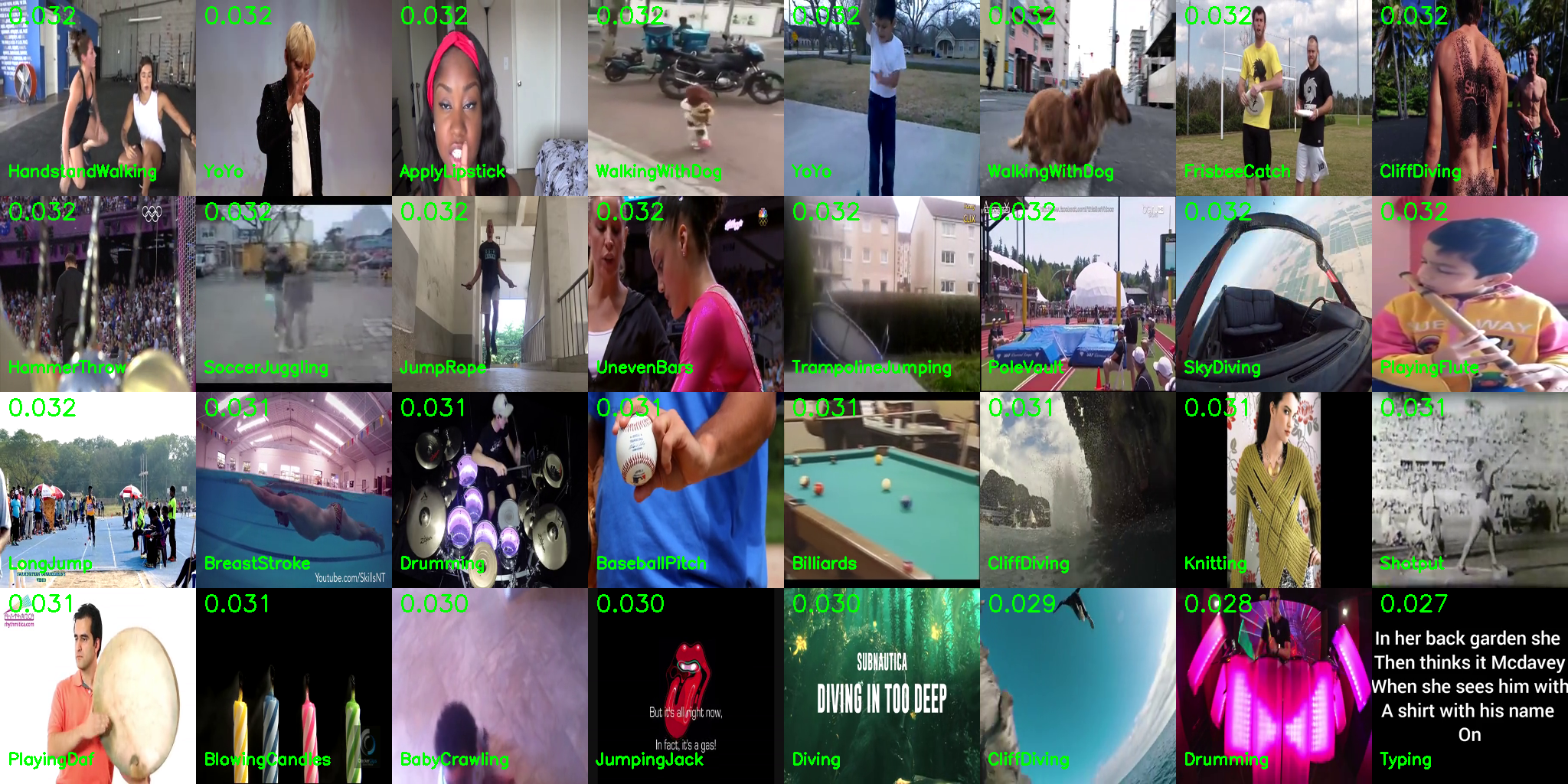}
\vspace{-0.5cm}
\caption{\textbf{Attention Weighting.}  For a web image batch, we show the weights ($\alpha$ in Equation~\ref{eq:attention}) for each image and the category of the image (the weights sum to 1). 
Images with lower weight in the last row tend to be more cartoon-like or contain excessive text while images with higher weight tend to be more representative of the action.
Images such as ``CliffDiving'' and ``Drumming'' in the last row appear reasonable and can be considered failure cases since they are assigned lower weight.
}
\label{fig:attention}
\end{figure*}

Next, we show an example of the weights learned by attention using a batch size of 32 in Figure~\ref{fig:attention}.  The weight ($\alpha$ in Equation~\ref{eq:attention}) is shown for each web image along with the category of the image (the weights sum to 1).  
Images that are cartoon-like or contain excessive text tend to receive lower weight since they appear less similar to images from the target domain (UCF-101~\cite{ucf-101}).  
A failure case of the attention component can be observed for the ``CliffDiving'' and ``Drumming''  images in the last row.  These images look reasonable but may have received lower batch score due to the extreme perspective and odd color palette, respectively.
%
Also attention does not help for images with the wrong semantic category (e.g. ``CliffDiving'' in the first row).

Table~\ref{tab:UCF-101-ablation} shows ablation study results on UCF-101~\cite{ucf-101}. For each row, the accuracy is averaged over the 3 splits of UCF-101.  
\begin{table}[t]
\centering
  \begin{tabular}{| c | c | c | c |} 
    \hline
    \textbf{Input} & \textbf{Arch} & \textbf{Features} & \textbf{Accuracy (\%)} \\ [0.5ex] 
    \hline\hline
    I & S & App & 62.984 \\
    \hline
    F & S &  App & 57.955 \\
    \hline
    I + F & S &  App & 70.117 \\ 
    \hline
    I + F (A) & T &  App & 71.365 \\
    \hline
    I + F (DA) & T &  App & 72.233 \\
    \hline
    I + F (A + DA) & T & App & 72.608 \\
    \hline
    \hline
    V & S & App + Temp & 72.563 \\
    \hline
    V (A) & D & App + Temp & 74.012 \\
    \hline
    V (DA) & D & App + Temp & 74.288 \\
    \hline
    V (A + DA) & D &  App + Temp & \textbf{74.876} \\
    \hline
  \end{tabular} 
\vspace{-0.3cm}
\caption{\textbf{Ablation study on UCF-101}. We evaluate the image and video models as well as the DA and attention components. We show the top-1\% performance of each model averaged over 3 splits of UCF-101~\cite{ucf-101}. 
Abbreviations are I: web image, F: web video frame, A: attention component, DA: domain adaptation component, V: web video, S: single branch (standard 2D CNN), T: triplet branch, D: dual branch (Siamese), App: appearance, Temp: temporal.
%
}
\label{tab:UCF-101-ablation}
\end{table}
\begin{table}
\centering
  \begin{tabular}{| c | c | c | c |} 
    \hline
    \textbf{Input} & \textbf{Arch} & \textbf{Features} & \textbf{Accuracy (\%)} \\ [0.5ex] 
    \hline\hline
    I + F & S & App & 39.632 \\ 
    \hline
    I + F (A) & T & App & 41.808 \\
    \hline
    I + F (DA) & T & App & 41.946 \\
    \hline
    I + F (A + DA) & T & App & 42.263 \\
    \hline
    V & S & App + Temp & 42.220 \\
    \hline
    V (A) & D & App + Temp & 42.527 \\
    \hline
    V (DA) & D & App + Temp & 42.506 \\
    \hline
    V (A + DA) & D & App + Temp & \textbf{42.817} \\
    \hline
  \end{tabular}
  \vspace{-0.3cm}
\caption{\textbf{Ablation study on Kinetics.}. We evaluate different model components and show the top-1\% accuracy of each model.  
The abbreviations are the same as in Table~\ref{tab:UCF-101-ablation}.
%
}
\label{tab:kinetics-ablation}
\end{table}
The inputs correspond to \textbf{I}: web images, \textbf{F}: web video frames, \textbf{I + F}: web images and video frames together, \textbf{V}: web video chunks.
%
In addition, we train on the different model components \textbf{A}: the attention component,  \textbf{DA}: the domain adaptation (adversarial) component, \textbf{A + DA}: both components.
%
The model architectures correspond to \textbf{S}: single branch (i.e. a standard 2D CNN), \textbf{D}: dual branch (i.e. a Siamese network) corresponding to the image model, \textbf{T}: triplet branch corresponding to the video model.
%
Lastly the features correspond to \textbf{App}: appearance (image) features,  \textbf{Temp}: temporal (video) features, \textbf{App + Temp}: both appearance and temporal features.

We can see that model a trained with images and video frames together (I+F) outperforms a model trained with image (I) and video frames (F) separately.  We verified that simply adding more images or video frames did not improve performance.
Next, we can see that adding the domain adaptation (DA) and attention (A) components separately helps improve performance by a small amount.  Adding both components leads to the best image model, I+F(A+DA), at 72.61\% top-1 accuracy.  

The next step is to take the spatial weights of the image model, initialize the video model and then continue training on web videos.
The video model V has an accuracy of 72.56\%, which is slightly lower compared to the image model accuracy of 72.61\%.  This may be due to irrelevant frames and noise present in the webly videos that are unaccounted for.  
Adding both the domain adaptation and attention components leads to our best performance of 74.88\% top-1 accuracy on UCF-101 for the video model V(A+DA).

We also explored a couple variations of training the video model V.
We first initialized V from ImageNet~\cite{ImageNet} spatial weights rather than the image model, which resulted in an accuracy of 59.12\%.  This drop in performance compared to model V (from 72.56\% to 59.12\%) may vindicate our two-step procedure of training an image model based on web images first, since training on web videos directly led to worse performance.  
%
In addition, we explored a variation of the video model V which \textit{does not} use the residual connection (corresponding to Figure~\ref{fig:spatiotemporal_block}a).  This model achieves an accuracy of 70.27\% which is still lower than V which got 72.13\%, which may indicate that adding the residual connection leads to an easier optimization.

We compare our approach to previous work on UCF-101 in Table~\ref{tab:UCF-101-other}.  
Among \ws approaches, we are competitive with the state-of-the-art LeadExceed~\cite{leadExceed} model at 76.3\%, vs 74.9\% for our model.  
%
%
LeadExceed requires 5 stages of model training/refinement steps, while our model unifies classification and filtering and requires only 2 stages.  Our model simplifies the training procedure at the cost of a small drop in accuracy (about 1.4\%).  
%
%
We note there is still a large gap between \ws methods and the state-of-the-art methods which directly use the UCF training data. 

We also evaluate on the larger Kinetics~\cite{kinetics} dataset. The results in Table~\ref{tab:kinetics-ablation} show similar improvements from adding the attention and domain adaptation components, leading to the best performance of 42.8\% accuracy.  We also compare against leading methods in Table~\ref{tab:kinetics} and note that there is a large gap between our \ws approach and state-of-the-art.  We are not aware of other \ws approaches evaluated on Kinetics.



\begin{table}
\begin{center}
  \begin{tabular}{| c | c | c | c | c | c |} 
    \hline
    \textbf{Approach} & \textbf{Type} & \textbf{Pre} & \textbf{Train} & \textbf{Acc(\%)} \\ [0.5ex] 
    \hline\hline
    UnAtt \cite{attentionTransfer} & App & IN & Web & 66.4 \\
    Webly \cite{webly} & App & IN & Web & 69.3 \\
    LeadExceed \cite{leadExceed} & App + Temp & IN & Web & 76.3\\
    Our model & App + Temp & IN & Web & 74.9 \\
    \hline
    C3D \cite{c3d} & App + Temp & K & UCF  & 82.3 \\
    2S \cite{twoStream} & App + Temp & IN & UCF & 88.0 \\
    R2D-2S \cite{r2d} & App + Temp & K & UCF & 97.3 \\
    I3D-2S \cite{quoVadis} & App + Temp & IN+K & UCF & \textbf{98.0} \\
    \hline
  \end{tabular}
\end{center}
\vspace{-0.5cm}
\caption{\textbf{UCF-101 Results.}  Comparison to several top approaches on UCF-101~\cite{ucf-101}. 
Abbreviations are App: appearance, Temp: temporal, Pre: pretraining data, Train: training data, Acc: top-1 accuracy, IN: ImageNet, K: Kinetics.
%
}
\label{tab:UCF-101-other}
\end{table}

\begin{table}
\begin{center}
\begin{tabular}{| c | c | c | c |} 
  \hline
  \textbf{Approach} & \textbf{Pretrained} & \textbf{Training} & \textbf{Acc(\%)} \\ [0.5ex] 
  \hline\hline
  C3D\cite{c3d} & ImageNet  & Kinetics & 57.0 \\
  \hline
  2S \cite{twoStream} & ImageNet & Kinetics & 61.0 \\
  \hline
  R2D-RGB~\cite{r2d} 2S &  Sports-1M & Kinetics & 75.4 \\
  \hline
  I3D-2S\cite{quoVadis} & ImageNet & Kinetics & 75.7 \\
  \hline
  NL I3D~\cite{nonLocalNN} & ImageNet & Kinetics & 77.7 \\
  \hline
  SlowFast~\cite{slowFast} & None & Kinetics & \textbf{79.8} \\
  \hline
  Our model & ImageNet & Web & 42.8 \\
  \hline
\end{tabular}
\end{center}
\vspace{-0.5cm}
\caption{\textbf{Kinetics-400 Results.}  Comparison to popular approaches on Kinetics~\cite{kinetics} for top-1\% accuracy on the validation set. 
The Two-Stream model is abbreviated as 2S.
%
}
\label{tab:kinetics}
\end{table}

\section{Conclusion}
We have presented a new model for video classification using only \ws data.  
Our model proceeds in two stages by first learning an image model, transferring the spatial weights to the video model, and continuing training with videos.
Our model also incorporates an adversarial component to learn a domain-invariant feature representation between source and target domains and accounts for noise using a novel attention component.
We demonstrated performance competitive with state-of-the-art for \ws approaches on UCF-101~\cite{ucf-101} while simplifying the training procedure, and also evaluated on the larger Kinetics~\cite{kinetics} for comparison.

{\small
\bibliographystyle{ieee}
\bibliography{bibliography}
}

\end{document}